\documentclass[letterpaper]{article} % DO NOT CHANGE THIS
\usepackage[preprint]{aaai2027}  % DO NOT CHANGE THIS

\providecommand{\say}[1]{``#1''}

\usepackage[hyphens]{url}  % DO NOT CHANGE THIS
\usepackage{graphicx} % DO NOT CHANGE THIS
\usepackage{natbib}  % DO NOT CHANGE THIS AND DO NOT ADD ANY OPTIONS TO IT
\usepackage{caption} % DO NOT CHANGE THIS AND DO NOT ADD ANY OPTIONS TO IT
\usepackage{algorithm}
\usepackage{algorithmic}

\usepackage{newfloat}
\usepackage{listings}
\DeclareCaptionStyle{ruled}{labelfont=normalfont,labelsep=colon,strut=off} % DO NOT CHANGE THIS
\floatstyle{ruled}
\newfloat{listing}{tb}{lst}{}
\floatname{listing}{Listing}

\usepackage{booktabs}
\usepackage{amsmath}
\usepackage{amssymb}

\title{\textsc{Orca}: Neural Operators for Causal Reasoning in Continuous Time}

\author{
    Gerrit Gro{\ss}mann\textsuperscript{\rm 1}\corresponding,
    David A. Selby\textsuperscript{\rm 1},
    Sebastian Vollmer\textsuperscript{\rm 1}
}
\affiliations{
    \textsuperscript{\rm 1}German Research Center for Artificial Intelligence (DFKI), Kaiserslautern, Germany\\
    Gerrit.Grossmann@dfki.de,
    David.Selby@dfki.de,
    Sebastian.Vollmer@dfki.de
}

\begin{document}

\maketitle

\begin{abstract}
Structural causal models are the standard language for reasoning about interventions and counterfactuals, but they describe static variables, typically measured once, and usually forbid cyclic dependencies. Many systems we care about, such as patients, climates, and economies, instead evolve continuously in time, are observed at irregular time points, and contain feedback loops. We argue that neural operator learning provides a natural foundation for causal reasoning in this setting, and propose \textsc{Orca}, a framework in which each node of the causal graph is a function of time and each mechanism is a learned map between function spaces. We extend existing neural operator architectures to express causal mechanisms: a mechanism computes the function value of a node from its parent nodes by taking several parent functions as input, respects the arrow of time, and treats latent exogenous noise as a function that can be inferred and reused for counterfactuals. We formalize the model class and demonstrate counterfactual reasoning on synthetic continuous-time examples.
\end{abstract}

\begin{links}
    \link{Code}{https://github.com/gerritgr/orca}
 \end{links}

\section{Introduction}

\begin{figure*}[t]
    \centering
    \includegraphics[width=0.6\textwidth]{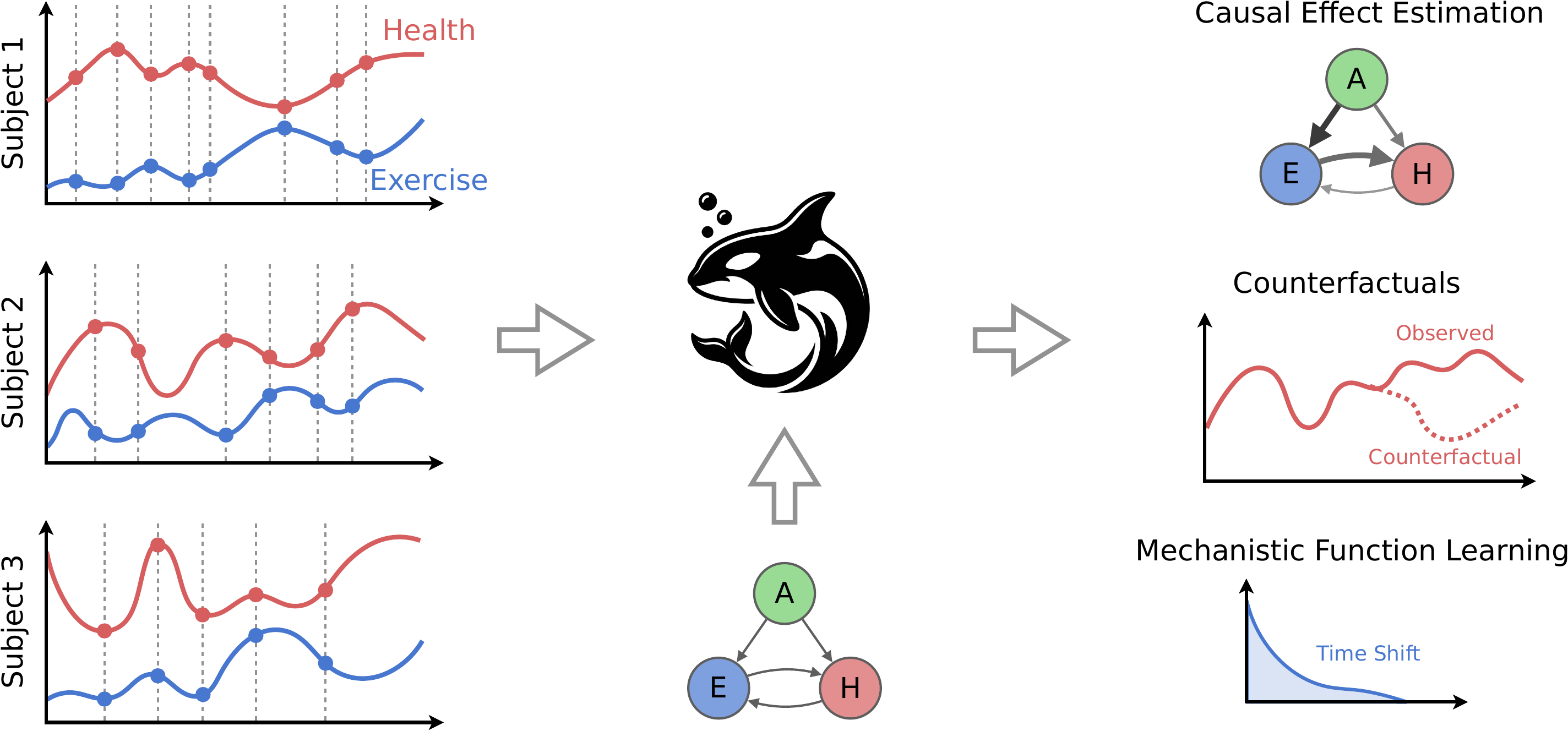}
    \caption{Schematic of \textsc{Orca} using the illustrative example of the age, exercise, and health relationship.
    \emph{Left}:
    Health and exercise trajectories are recorded for each subject at irregular time points.
    \emph{Center}: In addition, \textsc{Orca} receives the directed causal graph as input.
    It treats every node as a function of time and learns each structural equation as a neural operator that respects both the graph and the temporal priority principle.
    \emph{Right}: \textsc{Orca} equips this data modality with causal semantics. The trained model answers causal queries, including interventional causal-effect estimation, individual counterfactual trajectories, and recovery of the underlying mechanism, here a decaying exercise-effect kernel.}
    \label{fig:overview}
\end{figure*}

Causal questions are questions about what happens under change: \say{What would the average health be if everyone exercised this much?} or \say{What would this patient's health have been, had they exercised twice as much?}. Often, we want to reason about such questions even though only observational data are available, which is possible only under strong assumptions.

\paragraph{Causality.}
Within computer science and AI, the structural causal model (SCM) framework introduced by Judea \citet{pearl2009causality} has emerged as the standard formal language for causal reasoning and for formalizing such assumptions. Yet, the classical framework describes static variables without cyclic dependencies, while many systems we care about, such as patients, climates, economies, and epidemics, evolve continuously in time, are observed at irregular time points, and are often characterized by feedback loops \citep{runge2019inferring,christiansen2022toward}.
Extending SCMs and related causal frameworks to time-series data with cyclic dependencies is an active area of research. Existing approaches include unfolding SCMs over discrete time steps \citep{oberst2019counterfactual}, combining causality with dynamical systems analysis \citep{peters2022causal,palu2018causality-5ee,ness2019integrating-ecb,aalaila2026chaotic}, and measure-theoretic axiomatizations that incorporate both time and cycles \citep{park2023measure}.

\paragraph{Neural Operator Learning.}
We argue that neural operators provide a natural basis for causal reasoning about continuous data that are sampled at potentially irregular discrete time points, while also allowing (non-instantaneous) cyclic dependencies.
Neural operators are a class of neural networks designed to learn \textit{discretization-invariant} mappings between function spaces. That is, they can be trained and evaluated on different discretizations of the underlying continuous functions, rather than being tied to a fixed sampling grid. In suitable settings, refining the discretization yields outputs that converge to the corresponding continuous operator. Different implementations have been proposed, including graph neural operators \citep{li2020neural}, Fourier neural operators \citep{li2021fourier}, and DeepONets \citep{lu2021learning}. General recipes for constructing such neural networks are emerging \citep{berner2026principled}.

\paragraph{Contribution.}
This paper presents \textsc{Orca} (Operator-based Reasoning for Causal Analysis), a framework for causal reasoning built on the principles of neural operator learning (Figure~\ref{fig:overview}).
Specifically, \textsc{Orca} equips continuous-time data (sampled at arbitrary timepoints) with a causal semantics that supports causal effect estimation, counterfactual reasoning, and the learning of mechanistic functions, which we demonstrate on several synthetic examples.

\textsc{Orca} uses a directed causal graph in which each node represents a continuous function of time (e.g., blood oxygen). The mechanistic functions that determine the value of a node from its parents are expressed as learned maps between function spaces. Unlike traditional function learning with neural operators, which typically learns a single mapping from one input function to one output function \citep{shi2026stochastic}, \textsc{Orca} (i) allows for multiple input functions, one for each parent in the causal graph, (ii) enforces the \textit{temporal priority principle} (or \say{arrow of time}) which ensures that a cause must precede its effect, and (iii) supports the inference of latent noise variables, which are needed to compute counterfactuals.

\paragraph{Structure.}
The next section develops the argument using a running example, first in the static setting suited to structural causal models, then in continuous time where they break down. The following section formalizes the underlying model class and shows why neural operators are its natural instantiation. We then summarize implementation details, limitations, and conclusions.

\paragraph{Related Work.}
Causal semantics for dynamical systems have been developed for ordinary and stochastic differential equations, dynamic structural causal models, and measure-theoretic axiomatizations \citep{bongers2018causal, peters2022causal, sokol2014causal,park2023measure, boeken2024dynamic, boeken2026causal}. These works supply causal semantics but no learning recipe. Conversely, neural operators \citep{kovachki2023neural, berner2026principled} and their generative extensions \citep{shi2024universal, shi2026stochastic} learn maps and distributions over function spaces, but expose no interventional or counterfactual semantics. Causal inference for time series, in turn, commonly discretizes time, which ties its conclusions to the sampling grid \citep{runge2019inferring, reisach2025case}. \textsc{Orca} instantiates continuous-time causal semantics with learned, resolution-invariant operators.

\section{Illustrative Example}
Our running example has three variables: age $A$, exercise $E$, and health $H$, with causal graph $A \to E$, $A \to H$, and $E \to H$. Age is a confounder, because older subjects tend to exercise less and tend to be less healthy. Later, we also add a cyclic dependency, in which people in poor health exercise less.

\begin{figure}[t]
\centering
\includegraphics[width=0.9\columnwidth]{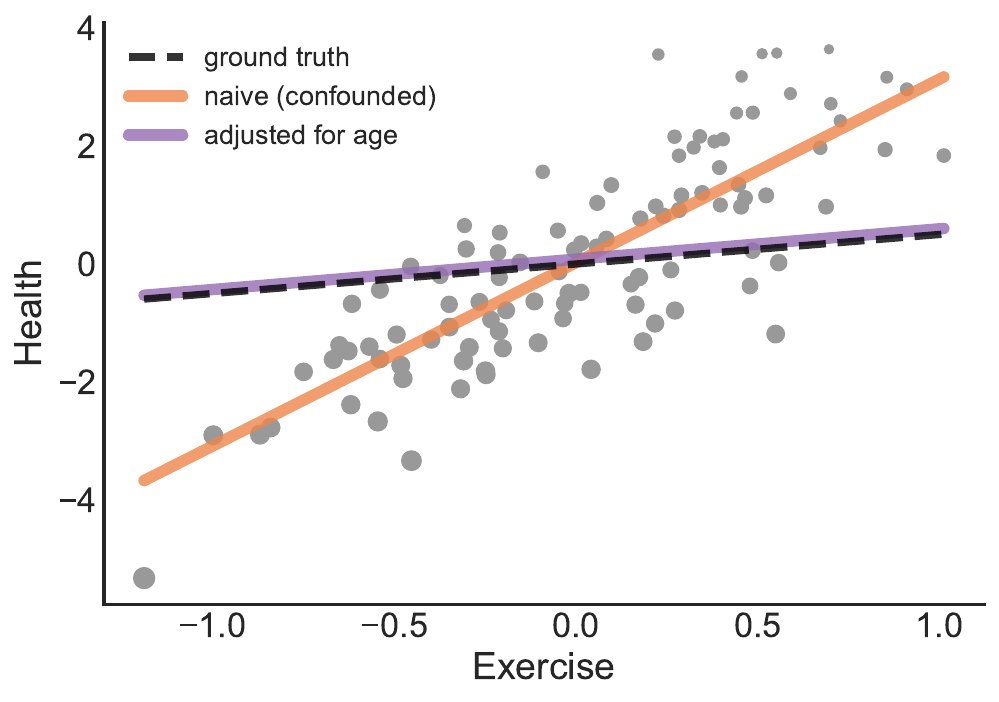}
%\fbox{\parbox[c][4cm][c]{0.9\columnwidth}{\centering\itshape Placeholder: static SCM figure.}}
\caption{Static example with one measurement of age, exercise, and health per subject. Points show the observed data. The naive dose--response estimate, which regresses health on exercise alone, is confounded by age and overestimates the effect of exercise. The causally informed estimate adjusts for age and recovers the ground-truth dose--response curve. The raw data is shown as gray dots, size indicates age. }
\label{fig:scm_static}
\end{figure}

\paragraph{Why Structural Causal Models Are Useful.}
Assume we have data for 100 subjects, with one measurement of age, exercise, and health for each subject at some point in their lives.
An SCM provides a structural equation to compute each variable from its causal parents and an exogenous noise term. For example,
\begin{equation*}
\begin{aligned}
E &= -0.5A + N_E, \\
H &= -2A + 0.5E + N_H .
\end{aligned}
\end{equation*}
Here $N_E$ and $N_H$ represent unobserved Gaussian noise.
In the real world, we may have the data and the causal graph, but not the equations. Three causal questions arise. The first is how to estimate the equation for $H$, parametrically or nonparametrically. The second is how much exercise influences health, formalized as a dose--response curve. Importantly, this is not merely the correlation between exercise and health: older people tend to exercise less \emph{and} be in worse health, so a correlation appears even without a causal mechanism. The dose--response curve instead asks how health would change if we \emph{forced} a subject to exercise a specific amount, written $\mathrm{do}(E = e)$ \citep{pearl2009causality}. It can be computed from the graph and the equation for $H$, or inferred from data alone, by comparing subjects of the same age but different exercise levels, using the \emph{backdoor} criterion, and then averaging over the population. The third question is counterfactual: what would this specific person's health have been, had they exercised more? Counterfactuals assume that the noise encodes subject-specific latent values that can be estimated and reused. We return to them below. Figure~\ref{fig:scm_static} shows the observed data together with the na\"ive, confounded dose--response estimate and the causally adjusted estimate.

\paragraph{Static Causal Models Are Not Enough for Time Series.}
In reality, exercise and health are not single numbers, but processes that unfold over decades. Longitudinal data also arrive at irregular, subject-specific time points.
A classical SCM may break in this setting or hide relevant assumptions. It cannot express in a principled way that health status today depends on exercise \emph{accumulated over the past}. It also struggles with cyclic dependencies, which are ubiquitous in real-world systems, e.g.\ when poor health reduces future exercise. Finally, any fixed discretization, such as one value per year, ties the model and its causal conclusions to an arbitrary measurement grid.
Observing a subject at one resolution and then computing a counterfactual trajectory at a different resolution is not possible in a principled way.

\paragraph{The Same Example in Continuous Time.}
We therefore lift the example to continuous time (Figure~\ref{fig:overview}): each node $A_i(t)$, $E_i(t)$, $H_i(t)$ becomes a function $[0, T] \rightarrow \mathbb{R}$, and we add a cyclic edge $H \to E$, representing the idea that people in poor health exercise less. Each node carries input noise, modeled here through Brownian motion. For each of $n = 100$ subjects, let $A_i(0)$, the age of the subject at the start of follow-up (in years), be drawn uniformly from $[10, 50]$ and assume $A_i(t)$ grows linearly. Then exercise level follows
\begin{equation}
E_i(t) = c / A_i(t) + \gamma\, H_i(t - \Delta) + B^E_i(t),
\end{equation}
where $c = 80$, $\gamma = 0.2$ couples exercise to past health at a delay of $\Delta = 1$ year (with $H_i(s) = H_i(0)$ for $s \le 0$), and $B^E_i$ is a Brownian motion. Health accumulates the effect of past exercise through a convolution kernel,
\begin{equation}
\small
H_i(t) = H_i(0) - 0.15\, A_i(t) + B^H_i(t) + \int_0^{t} \alpha e^{-\lambda (t-s)} E_i(s)\,\mathrm{d}s ,
\label{eq:truemech}
\end{equation}
with $\alpha = 0.3$, $\lambda = 0.2$, baseline health $H_i(0) \sim \mathcal{N}(5, 1)$, and Brownian motion $B^H_i$. We simulate $T=20$ years and sample between $10$ and $100$ irregular measurement points per subject, all variables measured at once and the first measurement always at $t = 0$. The \emph{dose--response} query takes as input some value $e$ and sets $\mathrm{do}(E = e)$ during the first ten years. It outputs health two years later, tracing out $\mathbb{E}\big[H_i(12) \mid \mathrm{do}(E = e)\big]$.  

\begin{figure}[t]
\centering
 \includegraphics[width=0.9\columnwidth]{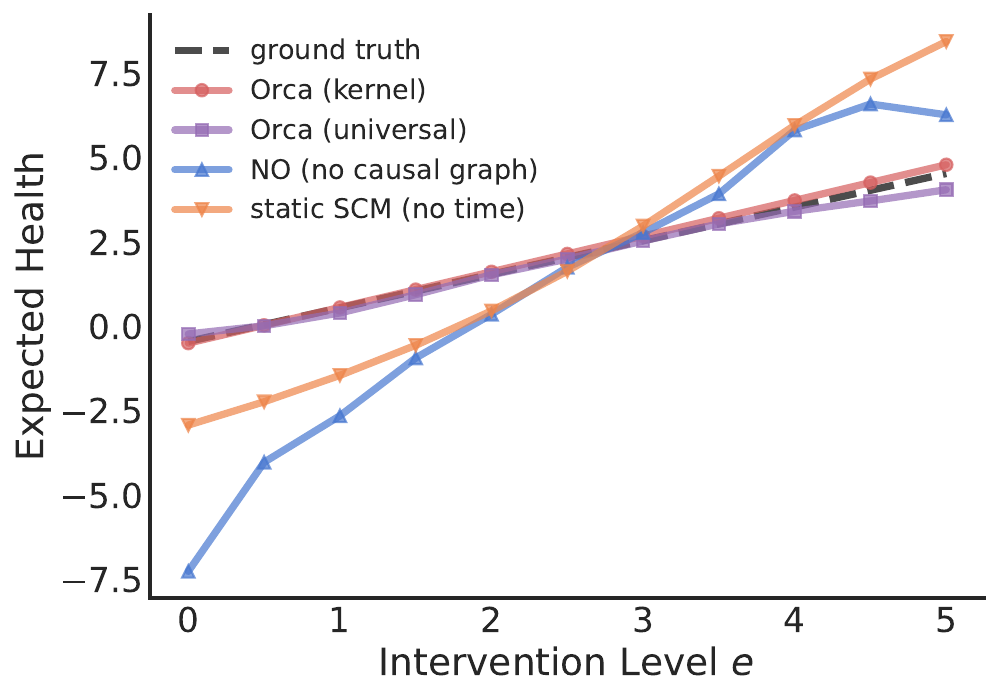}
%\fbox{\parbox[c][4cm][c]{0.9\columnwidth}{\centering\itshape Placeholder: dose--response comparison.}}
\caption{Continuous-time dose--response curves $\mathbb{E}[H_i(12) \mid \mathrm{do}(E = e)]$. The ground truth is compared with two naive estimators, a static structural causal model that ignores time and a neural operator that ignores the causal graph (no age), as well as with the two \textsc{Orca} variants (kernel operator and universal operator). Both \textsc{Orca} variants track the ground truth, while the static model and the causally blind operator are biased.}
\label{fig:no_dose}
\end{figure}

To compute the ground truth (Figure~\ref{fig:no_dose}), we reuse the same cohort of subjects: each subject's age and baseline health stay fixed, while we re-draw fresh Brownian noise and re-simulate their exercise and health using the model where $E$ is set to the input value $e$. Averaging the resulting health at year 12 over subjects and noise draws gives the curve.

\paragraph{Two Naive Estimators Fail in Characteristic Ways.}
First we illustrate two failure cases. 
(i) \emph{Ignoring the causal graph}: a neural operator trained to predict the health trajectory directly from the exercise trajectory, without age, is confounded just like the naive regression above and overestimates the dose--response slope. (ii) \emph{Ignoring time}: a static SCM that treats every measurement tuple $(a, e, h)$ as an independent sample adjusts for age but cannot represent the accumulation of exercise, and its dose--response curve is biased everywhere. Both deviate clearly from the ground truth (Figure~\ref{fig:no_dose}).

\paragraph{Using a Causally Informed Neural Operator.}
\textsc{Orca} instead learns the mechanism in Equation~\ref{eq:truemech} as an operator that respects the causal graph, the temporal priority principle, and the \emph{structure} of the mechanism. 
In the simplest case, we assume we know that a kernel acts on exercise alone and age is incorporated afterwards. 
Taking each subject's first measurement as the time origin, so that $H_i(0)$ is observed rather than latent, the model is
\begin{equation}
\begin{aligned}
I_i(t)
&= \int_{0}^{t}
K_{\theta_1}(t - s)\, E_i(s)\, \mathrm{d}s, \\
\hat H_i(t)
&= H_i(0) + I_i(t)
  + \rho_{\theta_2}\big(A_i(t), t\big),
  \quad t > 0 ,
\end{aligned}
\label{eq:kernelno}
\end{equation}
with small multilayer perceptrons $K_{\theta_1}$ and $\rho_{\theta_2}$. The head $\rho_{\theta_2}$ adjusts the kernel contribution additively, which pins the scale of the learned kernel. We fit $\theta_1$ and $\theta_2$ by minimizing a mean squared error over the observations, weighted by the local measurement spacing so that densely sampled subjects do not dominate. The next section develops this resolution-invariant training in general. Because $K_{\theta_1}(\tau)$ and the true kernel are the same kind of object, a function of the time lag only, they can be compared directly. The learned kernel is interpretable as the effect of exercise performed $\tau$ years ago and matches the truth closely for longer lags, while the very recent past is only weakly identified by sparse measurements. 

To estimate the dose--response curve from the learned operator (Figure~\ref{fig:no_dose}), we repeat the simulation with the true health mechanism replaced by the trained operator. For each level $e$ we again fix each subject's age and baseline health, draw fresh noise from the noise model, and simulate the intervened cyclic system forward: exercise is held at $\mathrm{do}(E = e)$ during the first ten years and afterwards follows its own mechanism (assumed known here, in general learned the same way) driven by the operator's health predictions, while health is produced by the operator. Averaging the predicted health at year twelve over subjects and noise draws gives the curve. 
%Importantly, this uses only samples from the noise model, never a subject's actual noise, so no abduction is required. 
%The result closely follows the ground truth and clearly separates from the two naive estimators (Figure~\ref{fig:no_dose}).

\paragraph{Universal \textsc{Orca}.}
The kernel operator is, admittedly, somewhat restrictive: we already knew that the true mechanism is an exponential kernel acting on exercise with age added on top, and we built that form into the model. In general, the mechanism is unknown, and \textsc{Orca} learns it with a universal approximator, which we develop in the next section. This \emph{\textsc{Orca} universal operator} still tracks the ground truth (Figure~\ref{fig:no_dose}, purple squares).

\section{Our Method: \textsc{Orca}}
We now formalize the recipe of the example in the previous section.

\paragraph{The Generative Model.}
We use a cyclic continuous-time causal model that is solved over a time interval $[0, T]$. The model is a tuple $(G, {F_v}, {N_v})$. The graph $G$ has nodes $V = {1, \dots, n}$ and directed edges. Cycles are allowed. The set of parent nodes $\mathrm{pa}(v)$ is defined by the incoming edges of $G$. We assume an implicit self-loop, so that $v \in \mathrm{pa}(v)$.
Each node $v$ is equipped with an exogenous noise input $N_v$ implementing a stochastic process on $[0, T]$. The noise processes are sampled independently. For simplicity, we take them to be Brownian motion or similar stochastic processes. Random initial values can be added if desired. We use $N_v(t) \in \mathbb{R}$ to denote the noise input value of node $v$ at time $t$.
The value of a node $v$ is a real-valued function $x_v \colon [0, T] \to \mathbb{R}$. This value is determined by the node's mechanism, implemented by a forward operator $F_v$, which returns the node's value at a query time $t$ from the past of its parents and the current noise:
\begin{equation*}
\begin{aligned}
x_v(t)
&= F_v\big(H_v(t),\, N_v(t)\, ,t\big), \\
H_v(t)
&= \{(u, t', x_u(t')) \mid \,
u \in \mathrm{pa}(v),\ t' < t\}.
\end{aligned}
\label{eq:mechanism}
\end{equation*}
The \emph{history} $H_v(t)$ collects the value of every parent in $\mathrm{pa}(v)$, tagged with the parent $u$ it came from, at times strictly before $t$. This includes the node's own earlier values through the self-loop.
The noise enters separately, as its current value $N_v(t)$. Unlike the parents, whose values are read strictly before $t$, this value at $t$ is available.
Because every parent in $\mathrm{pa}(v)$ enters only for $t' < t$, the value at $t$ can depend only on the past. This gives the temporal priority principle and lets the self-loop act without instantaneous circularity.
When the history is empty, $F_v$ maps the noise alone to the value, which implicitly defines the node's initial distribution.
A useful special case is \emph{additive noise}, where the mechanism splits into a deterministic function of the past plus the current noise,
\begin{equation}
x_v(t) = F_v\big(H_v(t), t\big) + N_v(t),
\label{eq:additive}
\end{equation}
which we assume below because it lets us recover the noise from data as a residual.

\paragraph{Simulation.}
To draw a trajectory we fix a time grid $0 = t_0 < t_1 < \dots < t_J < T$ and first sample the exogenous inputs, namely the noise function $N_v$ of every node, on this grid; the noise is the model's only exogenous input. We then sweep forward in time: at each grid point $t_j$, and for every node $v$, we set $x_v(t_j) = F_v\big(H_v(t_j),\, N_v(t_j),\, t_j\big)$, where $H_v(t_j)$ gathers the already-computed parent values at earlier grid points $t_{j'} < t_j$ and the noise is its value at $t_j$. The first step uses an empty history and therefore draws the initial values from the noise alone. Because the parents in $\mathrm{pa}(v)$ are read strictly from the past, this forward sweep is well defined even though $G$ contains cycles: unrolled in time, the dependencies form a directed acyclic graph.

\paragraph{Semantics.}
The direct outcome of a single simulation is a discretized trajectory for each node. 
We turn this discrete solution into a continuous one by holding each value constant until the next grid point updates it, that is, $x_v(t) = x_v(t_j)$ for $t \in [t_j, t_{j+1})$, yielding a piecewise-constant, right-continuous trajectory. Thus, for any fixed time grid, the causal model induces a joint probability distribution over trajectory collections $(x_v)_{v \in V}$, where each simulation sample contains one trajectory for every node. We therefore view the semantics of the model as the limit of these grid-based distributions as the grid is refined, understood in the sense of weak convergence. The operator formalization makes this construction natural, but it does not remove the need for regularity assumptions on the mechanisms and noise processes. Avoiding pathological cases remains the modeler's responsibility.

\paragraph{Data and Objective.}
We assume that the input consists of irregular measurements of all nodes for a population of subjects (for simplicity, we assume that all nodes are measured at the same time points), and that the directed causal graph $G$ is known. The objective is to find a neural representation of the forward operators $F_v$ and, depending on the end goal, to also infer the latent input noise $N_v$. We can then use the learned $F_v$ and $N_v$ for downstream tasks such as computing dose--response curves, average treatment effects, or counterfactuals.

Formally, the input is a set $X = \{(i, u, t, x_{i,u}(t))\}$, where $i$ is a subject indicator, $u$ is a node indicator, $t$ is the measurement time, and $x_{i,u}(t)$ is the value of node $u$ of subject $i$ at $t$.

\paragraph{Training and Architecture.}
To learn the mechanism of a node $v$, we iterate over all measured tuples $(i, v, t, x_{i,v}(t))$ and predict $x_{i,v}(t)$ from the part of $X$ that the mechanism is allowed to see, namely the same subject's measurements of $v$'s parents $\mathrm{pa}(v)$ (which, by the self-loop, includes $v$ itself) taken before $t$,
\begin{equation*}
S_v^{\,i}(t) = \{\, (i, u, t', x_{i,u}(t')) \in X \mid \, u \in \mathrm{pa}(v),\ t' < t \,\}.
\end{equation*}

Under the additive-noise assumption of Equation~\ref{eq:additive}, the noise is just the residual, so it need not be an input to the learned mechanism: we fit the deterministic part $f_v$ by minimizing prediction error alone. We learn it as a deep-set operator with parametrization $\theta = (\theta_1, \theta_2)$: a shared embedding network $\phi_{\theta_1}$ and a decoding network $\rho_{\theta_2}$. To predict $x_{i,v}(t)$, we embed every measurement in $S_v^{\,i}(t)$ with $\phi_{\theta_1}$, sum the embeddings, and decode the result together with the query time $t$ using $\rho_{\theta_2}$,
\begin{equation*}
\hat x_{i,v}(t) = \rho_{\theta_2}\Big(t,\ \sum_{(i, u, t', x_{i,u}(t')) \in S_v^{\,i}(t)} w_{t'}\, \phi_{\theta_1}\big(u, t', x_{i,u}(t')\big)\Big).
\end{equation*}

The weight $w_{t'}$ turns the sum into an integral estimate. For each subject $i$ and parent $u$ we sort the measurement times $t'_1 < t'_2 < \dots$ and set $w_{t'_k} = \tfrac{1}{2}(t'_{k+1} - t'_{k-1})$, the span of time that measurement $t'_k$ stands for (with one-sided gaps at the first and last point). Without this weighting the sum would simply grow with the number of measurements; with it, the contribution of each parent $u$ approximates $\int_0^{t} \phi_{\theta_1}\big(u, s, x_{i,u}(s)\big)\,\mathrm{d}s$, the integral of the embedding along that parent's trajectory, so the result no longer depends on how densely we sample. Feeding only the parents enforces graph consistency, and restricting $S_v^{\,i}(t)$ to earlier measurements ($t' < t$) enforces temporal priority.

Let $\mathcal{T}_{i,v} = \{\, t \mid (i, v, t, x_{i,v}(t)) \in X \,\}$ be the measurement times of node $v$ for subject $i$. We fit the parametrization $\theta$ by gradient descent on the subject-normalized mean squared error
\begin{equation}
\mathcal{L}_v(\theta)
=
\frac{1}{M}
\sum_{i=1}^{M}
\frac{1}{|\mathcal{T}_{i,v}|}
\sum_{t \in \mathcal{T}_{i,v}}
\big(\hat x_{i,v}(t) - x_{i,v}(t)\big)^2 .
\label{eq:training_loss}
\end{equation}
The factor $1/|\mathcal{T}_{i,v}|$ makes densely and sparsely sampled subjects contribute equally to the overall loss.

\paragraph{Resolution Consistency.}
The architecture is motivated by two established results. First, the mechanism we want to learn is a continuous, permutation-invariant functional of the parent measurements, and the DeepSets representation theorem motivates the encode-sum-decode form
$\rho_{\theta_2}\!\big(t, \sum_k \phi_{\theta_1}(\cdot)\big)$
\citep{zaheer2017deep}. MIONet provides a related universal-approximation result for operators with several input functions \citep{jin2022mionet}, supporting the use of multiple parent trajectories as inputs. Second, the parameters $\theta$ do not depend on the sampling resolution. For fixed $\theta$, the weighted sum
$\sum_k w_{t'_k}\,\phi_{\theta_1}(\cdot)$
is a quadrature approximation of
$\int_0^t \phi_{\theta_1}\big(u,s,x_{i,u}(s)\big)\,\mathrm{d}s$.
Under standard continuity assumptions, this approximation converges to the integral as the maximum spacing between measurements tends to zero. Predictions therefore converge under increasingly fine sampling, and the same learned model can be evaluated on sampling grids that differ from those used during training.

\paragraph{Interventional and Counterfactual Distributions.}
Once the mechanisms are learned, we answer causal queries by simulating the intervened model, which requires three ingredients. First, a \emph{target}: a functional of the trajectories we care about, typically the value $x_{v^\ast}(t^\ast)$ of an outcome node $v^\ast$ at a readout time $t^\ast$. Second, an \emph{intervention}: replacing a node's mechanism on an interval, written $\mathrm{do}(x_v(t) = e,\ t \in [t_1, t_2])$, which fixes $x_v$ to the value $e$ there and cuts its dependence on its own parents, while all other mechanisms are left unchanged. Third, a way to \emph{compute} the query: we push the noise distribution through the intervened model, that is, we sample the noise processes, run the forward sweep with the modified mechanism, and read off the target, which yields the interventional distribution of $x_{v^\ast}(t^\ast)$ under $\mathrm{do}(x_v = e)$.
Two standard summaries follow. The \emph{dose--response curve} is the mean target as a function of the intervention level, $e \mapsto \mathbb{E}\big[x_{v^\ast}(t^\ast) \mid \mathrm{do}(x_v = e)\big]$. The \emph{average treatment effect} is the difference in mean target between two interventions, $\mathbb{E}\big[x_{v^\ast}(t^\ast) \mid \mathrm{do}(x_v = e_1)\big] - \mathbb{E}\big[x_{v^\ast}(t^\ast) \mid \mathrm{do}(x_v = e_0)\big]$; taking the same difference conditional on a subject's covariates or observed history gives the conditional average treatment effect (CATE).
Counterfactual trajectories, which ask how one specific subject's trajectory would have looked under a different intervention, require an assumption on the noise, since we must first infer (abduct) that subject's latent noise. Under additive noise this is simply the residual between the observed trajectory and the model prediction, which we hold fixed while re-simulating under the intervention.

\paragraph{Assumptions.}
The causal interpretation of \textsc{Orca} requires a correct and causally sufficient graph, mechanisms that remain invariant under the considered interventions, and sufficient support for the parent histories and intervention levels of interest. We further assume that the model class can represent the relevant mechanisms. Individual counterfactuals additionally require a correctly specified noise model and an assumption that allows the subject-specific noise to be inferred, such as additivity or monotonicity.

\paragraph{Generalizations.}
The same recipe extends to vector-valued, categorical, and count-valued nodes, alternative noise models, and space-time fields. It can also use the full noise history $N_v|_{[0,t]}$ and supports simple uncertainty estimates through subsampling or resampling of the input measurements.

\begin{figure*}[!t]
\centering
\includegraphics[width=0.8\textwidth]{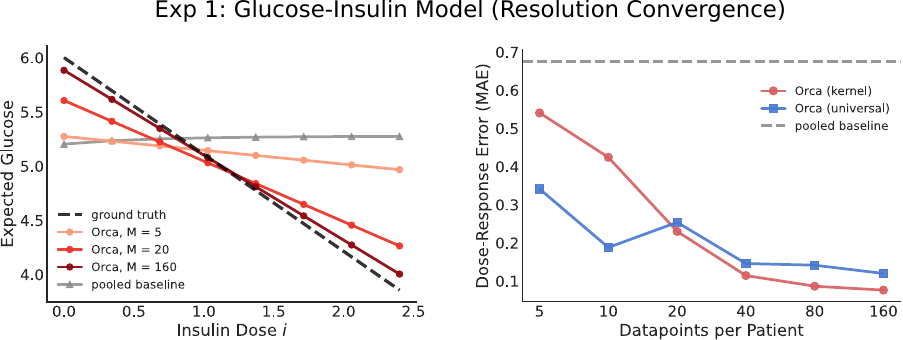}
\caption{Exp.\ 1, a glucose--insulin model in which a drifting insulin sensitivity confounds dosing and glucose, queried by the dose--response curve $\mathbb{E}[G(12) \mid \mathrm{do}(I = i)]$.
\textbf{Left}: the curve learned by the fixed-kernel operator at three sampling densities $M$, compared with the ground truth and a patient-agnostic pooled baseline.
\textbf{Right}: dose--response error against sampling density for both \textsc{Orca} variants. As the trajectories are observed more densely, both variants produce increasingly accurate estimates of the continuous-time dose--response curve, whereas the pooled baseline ignores the confounder and remains biased at every density.}
\label{fig:exp1}
\end{figure*}

\section{Learning Distributions over Functions}
So far we assumed additive noise (Equation~\ref{eq:additive}). Each mechanism splits into a deterministic part $F_v$ of the parent history plus the noise $N_v(t)$, so learning the mechanism reduces to a regression problem for $F_v$, and the noise can be recovered afterwards as a residual. Prediction is then conceptually a \emph{point estimate}, where the \say{point} is itself a function. For given parent trajectories, the model learns to predict a single best node function, namely the conditional mean.

In the general case, the noise is a genuine input to the mechanism,
$x_v(t) = F_v\big(H_v(t), N_v\big|_{[0,t]}, t\big)$,
and cannot be separated out. This is harder because $N_v$ is latent, so we must estimate $N_v$ and $F_v$ at the same time. The task therefore changes character. Instead of predicting a single \say{best} function, we predict a \emph{distribution over functions} for each node. Concretely, for each node of interest $v$, we learn how a particular input noise trajectory $N_v\colon [0,t] \rightarrow \mathbb{R}$ maps to a particular node trajectory $x_v\colon [0,t] \rightarrow \mathbb{R}$. This setting is much closer to stochastic process learning with neural operators \citep{shi2026stochastic}. Intuitively, the latent noise captures subject-specific factors. In the health example, we can read it as a form of \say{luck} that makes two subjects with identical histories differ. Unlike in the static case, this luck may change over time.

\paragraph{Assumptions.}
Giving up the strong additive-noise assumption adds computational cost, but it also lets us replace that assumption with new, and arguably weaker, ones. We assume that (i) we know a noise model, that is, a distribution over $N_v$ that reasonably captures the remaining uncertainty, and that (ii) we have a score function that says how compatible a sampled noise trajectory is with an observed node trajectory. Together, these assumptions let us build a coupling between sampled noise trajectories and observed subjects.

A simple way of constructing  a score function is based on the assumption of \emph{monotonicity}. We assume the mechanism is monotone in the noise, so that larger noise values tend to yield larger outcomes, conditional on the same parent history. In the static case, monotonicity is easy to use. It makes the noise equivalent to a rank, so each subject is summarized by a single number, namely where its outcome falls in the conditional distribution of outcomes. In our setting this is more complex, because the noise is a function of time rather than a single value. A subject may sit high in the distribution early in life and low later on, so its rank is not one number but a trajectory of ranks. Our formulation handles this by requiring only a score between a noise trajectory and a node trajectory. Such a score can express a time-varying rank, even if less directly than reading off a single conditional rank.

%\paragraph{Example.}
%Consider subject $0$. For the first ten years she is very lucky. Given her age and exercise level, she is consistently healthier than expected. We cannot observe this luck directly, because a clean comparison would require other subjects with exactly the same age and exercise history, which we never observe in practice. What we can do is estimate her expected health from the model and measure how far she deviates from it. Doing this, we find that in the first ten years she deviates upward more than any other subject, so she is the luckiest of all. In the second ten years the picture flips. She deviates downward more than anyone else, so she becomes the unluckiest. The noise trajectory that best represents her would therefore have to be higher than every other noise sample on $[0,10]$ and lower than every other sample on $[10,20]$. If we draw noise from a Brownian motion, we are unlikely to sample such an extreme trajectory by chance. There are two ways to deal with this. We could sample conditionally, so that we only keep noise trajectories consistent with the observations, or we could work with the noise samples we already have and map them to subjects by a heuristic. This work takes the second option. To give the mapping more room, we generate more noise samples than there are subjects, so that we can select a suitable subset rather than being forced to use every sample. This can be understood as a heuristic approximation to conditional sampling from the noise model.

\paragraph{Algorithm.}
Consider a single node of interest $v$ and $M$ subjects.
(i) We sample $K$ noise trajectories from the known noise model, with $K \geq M$.
(ii) We first train the model while ignoring the latent noise and compute residual trajectories as in the additive-noise case. We do not interpret these residuals as the noise itself. Instead, we use them as proxies for how each subject deviates from the conditional mean.
(iii) We then compute a matching between the residual trajectories and the sampled noise trajectories using an optimal transport objective. If $K$ is strictly larger than $M$, we select the best-matching noise trajectories and reject the others, which approximates conditioning the noise model on the observed subjects.
(iv) Once each subject has been coupled to a noise trajectory, all inputs to the mechanism are observed or assigned. We can then learn to predict the trajectory of node $v$ using the same deep-set architecture as before, now with the coupled noise trajectory included as an additional input.

The main difference from the additive-noise method is therefore that residuals over time are not interpreted as the noise itself. They are used only to couple observed subjects to plausible latent noise trajectories. The optimal transport objective is one possible coupling heuristic, but other choices are possible. For instance, one could place stronger weight on matching ranks at individual time points. A flow-matching approach could also be used to learn such a coupling, although direct autoregressive prediction is more natural when we want to enforce temporal priority.

This is the functional analogue of a standard recipe in static nonseparable structural models. Under monotonicity in the latent noise, the unobserved noise can be coupled to the observed outcome through its conditional rank, which allows one to replace additive residuals with rank-based latent variation \citep{matzkin2003nonparametric}. Similar rank-based ideas also underlie structural quantile models for heterogeneous treatment effects, where latent ranks are used to connect observed outcomes, counterfactual outcomes, and exogenous variation \citep{chernozhukov2005iv}.

\begin{figure*}[!t]
\centering
\includegraphics[width=0.85\textwidth]{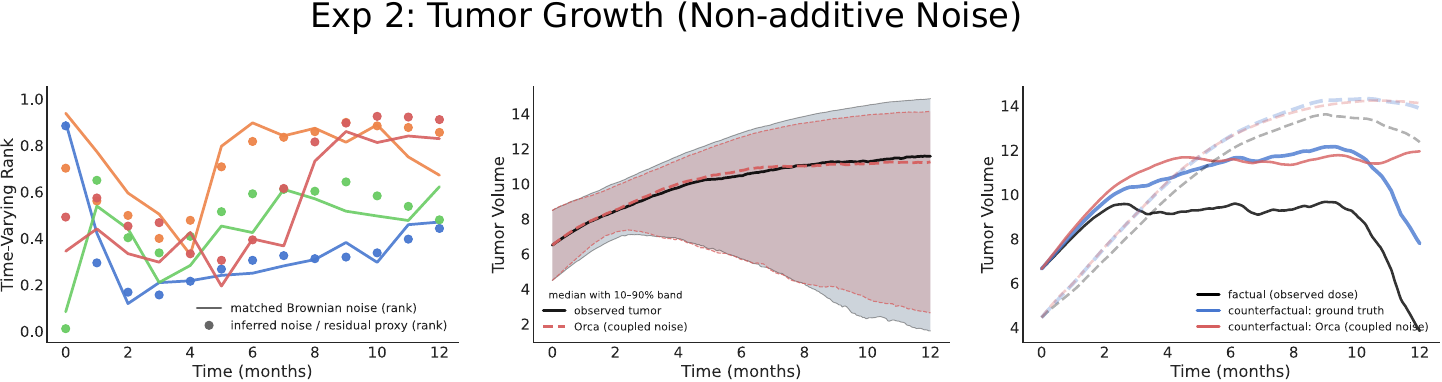}
\caption{Exp.\ 2, tumor growth under chemotherapy, where a latent, time-varying drug sensitivity multiplies the treatment effect, making the noise non-additive.
\textbf{Left}: the coupling step, for four selected subjects spanning the dose range. Markers show each subject's normalized rank at the discretization times, inferred from the residuals of a noise-free fit. Lines show the normalized rank of the Brownian sample that optimal transport coupled to that subject. The two line up closely.
\textbf{Center}: in black, the median tumor volume across all subjects, with the 10--90\% range in gray. We then re-simulate the entire population with the learned mechanism, driving each subject by its observed dose, its observed initial volume, and its coupled noise trajectory. This yields a second, synthetic population, whose median and 10--90\% range are shown in red. 
That the two coincide indicates that the learned mechanism and the coupled noise together reproduce the observed population, and in particular that the coupling carries genuine subject-level variability.
\emph{Right}: two individual counterfactuals, $\mathrm{do}(C \rightarrow C/2)$; Solid: the median-error subject. Dashed: the closest-agreeing high-dose subject. Both tumors regrow once the dose is halved.
}
\label{fig:exp2}
\end{figure*}

\section{Experiments}
Our experiments are small, controlled synthetic studies meant to demonstrate the properties and feasibility of the framework, not full-scale case studies. All are seeded and one-click reproducible, with the data-generating processes and training details in the Appendix and the accompanying code. Each defines a ground-truth continuous-time causal model, fits \textsc{Orca} from irregular samples, and compares its interventional and counterfactual estimates against the ground truth while varying one factor to probe a specific property.

\paragraph{Experiment 1: Glucose--Insulin (Resolution Convergence).}
This experiment tests whether the estimated causal conclusions converge as the measurement grid is refined. We sample the same underlying continuous-time process at different resolutions, from sparse to dense, and fit \textsc{Orca} separately at each resolution. The setup is a glucose--insulin model in which meals and insulin drive blood glucose, while dosing responds to recent glucose. As the observations become denser, the estimated dose--response curves converge toward the ground truth. Moreover, a model trained on sparse measurements can still answer queries on a finer grid than the one used during training (Figure~\ref{fig:exp1}).

\paragraph{Experiment 2: Tumor Growth (Counterfactuals under Non-additive Noise).}
This experiment tests the computation of individual counterfactuals when the noise is not additive, so that a subject's latent noise can no longer be recovered as a residual. The setup is a tumor-growth model under chemotherapy, whose response variability scales with tumor size. We couple observed subjects to sampled noise trajectories using a monotone rank-based optimal-transport procedure and reuse the inferred noise trajectories to compute individual counterfactuals (Figure~\ref{fig:exp2}).
Individual counterfactuals rest on assumptions that no amount of data can verify. \textbf{We therefore treat them as a diagnostic for the model class, not as claims about any real trajectory, and certainly not as a basis for real-world decisions}.

\section{Conclusions, Limitations, and Future Work}
We presented \textsc{Orca}, which uses neural operators as the natural mechanism class for causal models whose nodes are functions of time: they are resolution-consistent, handle irregular samples, and can be constrained to respect the causal graph and the temporal priority principle. On synthetic data, \textsc{Orca} recovered interventional dose--response curves and individual counterfactual trajectories, while baselines that drop either the causal or the temporal structure failed. The learned-kernel variant is the most interpretable and, when its structure matches the mechanism, the most accurate.

\paragraph{Limitations.}
We take the causal graph as given rather than learning it and assume causal sufficiency; unobserved common causes would bias the estimates. Even with the correct graph, predictive fit does not certify a causal decomposition: the parameterization must mirror the true mechanism, or causal estimates can be badly biased while the training loss barely moves. Counterfactuals further require an assumption on the noise, such as additivity or monotonicity, and the general non-additive case, which couples observed and sampled noise trajectories, is only sketched here. Because the node value integrates the noise, these residual proxies reflect accumulated rather than instantaneous deviations, overstating how persistent a subject's latent trajectory is. Finally, our data model is idealized (noise-free, synchronous measurements), and all validation is synthetic. Convergence under grid refinement is asymptotic, so under sparse sampling fast dynamics are weakly identified and long-horizon queries accumulate simulation error.

\paragraph{Future Work.}
Each limitation points at a natural next step: learning or validating the causal graph from irregular samples \citep{huang2020causal}, tolerating measurement noise and asynchronous observations, and turning the sketched coupling procedure for non-additive noise into a full method for learning distributions over functions (e.g., by combining causal structure with operator flow matching). Beyond that, the construction extends directly from time to space--time fields, and the decisive test will be applications to real longitudinal data, such as electronic health records.

\section{Reproducibility Statement}
Experiments are implemented in three self-contained notebooks with a fixed random seed on the Orca repository.

\bibliography{aaai2027}


\newpage
\appendix

\section{Appendix A: Experiment 1 in Detail}

This appendix describes the glucose--insulin experiment
(\texttt{notebooks/exp1\_data\_efficiency.ipynb}). Everything is synthetic and
generated inside the notebook.

\paragraph{The Simulated World.}
We simulate $1000$ patients over $24$ hours on a fine grid ($\Delta t = 0.05$).
Each patient has three time-varying quantities. Insulin sensitivity $X(t)$ is a
per-patient trait that drifts slowly; it starts at $X_i(0) \sim \mathcal{U}(0.3,
1.5)$ and wanders as a geometric Brownian motion. Insulin $I(t)$ is dosed in
response to recent glucose,
\begin{equation*}
I_i(t) = I_0 + m_i\, \kappa\, \frac{\max\big(G_i(t - \Delta) - G^\ast,\, 0\big)}{X_i(t)} + B^I_i(t),
\end{equation*}
with basal dose $I_0 = 0.6$, dosing gain $\kappa = 1.6$, target $G^\ast = 5.2$,
delay $\Delta = 0.5$\,h, a per-patient dosing style $m_i \sim \mathcal{U}(0.7,
1.4)$, and Brownian noise $B^I_i$. Glucose accumulates past insulin through an
exponential kernel, scaled by the patient's sensitivity,
\begin{equation*}
G_i(t) = G_0 - X_i(t) \int_0^t \gamma\, e^{-\lambda (t - s)} I_i(s)\, \mathrm{d}s + B^G_i(t),
\end{equation*}
with $G_0 = 6$, $\gamma = 0.6$, and $\lambda = 0.6$.

\paragraph{Why This Is a Causal Problem.}
Sensitivity $X$ is a confounder. A patient with low sensitivity receives
\emph{more} insulin \emph{and} has higher glucose. So insulin and glucose are
correlated even where insulin is doing its job, and any model that ignores $X$
will misjudge how much insulin actually helps. The graph also has a cycle:
glucose drives dosing, and dosing drives glucose.

\paragraph{What the Model Sees.}
Each patient is measured at $M$ irregular times, always including $t = 0$. All
three quantities are recorded at those times. The causal graph is given; the
equations are not.

\paragraph{The Causal Query.}
The dose--response curve $\mathbb{E}[G(12) \mid \mathrm{do}(I = i)]$: hold
insulin at a fixed level $i$ for the first $12$ hours and read glucose at hour
$12$, for eight levels $i \in [0, 2.4]$. The ground truth is obtained by
re-simulating the true system under the intervention with fresh noise. The same
noise draws are reused for every method, so differences between methods are not
noise artifacts.

\paragraph{Methods Compared.}
(i) A \emph{pooled baseline} that throws away patient identity and time, and
predicts glucose from the instantaneous insulin value alone. (ii) \emph{Orca
fixed-kernel}, which mirrors the mechanism: it learns the insulin kernel while
sensitivity is observed and enters multiplicatively. (iii) \emph{Orca universal},
the assumption-free deep-set operator, which embeds every past measurement, sums
the embeddings with quadrature weights, and decodes. Both Orca variants are
averaged over three random initializations.

\paragraph{What Is Varied.}
The sampling density $M \in \{5, 10, 20, 40, 80, 160\}$ points per patient. Every
model is refit from scratch at each density. We report the mean absolute error of
the estimated dose--response curve against the ground truth.

\paragraph{What the Figures Show.}
The first figure overlays the learned dose--response at three densities against
the truth and the pooled baseline. The second plots error against density. Both
\textsc{Orca} variants converge toward the truth as measurements become denser,
while the pooled baseline stays biased at every density: more data does not fix
a confounded model.

\section{Appendix B: Experiment 2 in Detail}

This appendix describes the tumor-growth experiment
(\texttt{notebooks/exp2\_tumor\_growth.ipynb}), which tests counterfactuals when
the latent noise is \emph{not} additive.

\paragraph{The Simulated World.}
We simulate $1000$ patients over $12$ months. Each patient has an observed
starting tumor volume $V_i(0) \sim \mathcal{U}(4, 9)$ and an observed constant
chemotherapy dose $C_i \sim \mathcal{U}(0.6, 1.3)$. The tumor grows and is killed
by the drug,
\begin{equation}
\frac{\mathrm{d}V_i}{\mathrm{d}t}
= \rho\, V_i \log\!\frac{K}{V_i}
- \beta\, e^{\sigma \varepsilon_i(t)}\, C_i\, V_i ,
\label{eq:tumor}
\end{equation}
with growth rate $\rho = 0.32$, carrying capacity $K = 16$, drug potency $\beta =
0.10$, and noise scale $\sigma = 0.55$. Volumes are floored at $0.1$, so a
near-complete response plateaus instead of hitting zero. The latent
$\varepsilon_i(t)$ is a Brownian path: the patient's drug sensitivity, drifting
over time. It is never observed.

\paragraph{Why the Noise Is Hard.}
In Equation~\ref{eq:tumor} the noise multiplies the drug term, so its effect
scales with both dose and current tumor size. The same $\varepsilon$ produces a
large swing in a large tumor and a small swing in a small one. Consequently the
noise cannot be read off as \say{observed minus predicted}: subtracting a
conditional mean does not isolate it. This is the non-additive case, and it is
what makes individual counterfactuals difficult, because a counterfactual
requires knowing \emph{this} patient's latent path.

\paragraph{What the Model Sees.}
Between $14$ and $30$ irregular measurements per patient, always including $t =
0$, recording tumor volume alongside the observed dose.

\paragraph{Recovering the Latent Noise.}
We follow the four-step recipe of the previous section.
(i) Sample $K = 2000$ Brownian trajectories from the known noise model, more than
the $1000$ patients, so poor candidates can be discarded.
(ii) Fit the operator once while ignoring the noise, and record each patient's
residual over time. These residuals are \emph{not} treated as the noise; they
only say how far a patient sits above or below the conditional mean.
(iii) Match patients to sampled trajectories by optimal transport. Because the
mechanism is monotone in the noise (more sensitivity means more kill means a
smaller tumor), both the residuals and the noise samples are converted to
\emph{ranks at each of $13$ time points}, and the transport cost is the squared
distance between rank trajectories. We solve the entropic problem with Sinkhorn
iterations and then extract a hard one-to-one assignment, rejecting the $1000$
unused samples. Matching on rank \emph{trajectories} rather than a single number
matters: a patient may respond well early and poorly later, and the assigned
noise path should have that same shape.
(iv) Refit the operator with the assigned noise as an input.

\paragraph{The Operator.}
As in the fixed-kernel variant of Experiment 1, the operator mirrors the
mechanism. The Gompertz growth law is assumed known in form; the model learns its
constants together with a monotone map from noise to drug sensitivity,
$s_\theta(\varepsilon) = \mathrm{softplus}(b + w\,\varepsilon)$, and integrates
the resulting equation as a parametric ODE with four learned constants. A counterfactual is then obtained
by re-integrating with a changed dose while the assigned noise is held fixed.
After fitting, the learned sensitivity slope is close to the true noise scale
($w \approx 0.5$ against $\sigma = 0.55$).

\paragraph{The Causal Query.}
An individual counterfactual: $\mathrm{do}(C \rightarrow 0.5\,C)$, that is,
\say{what would this patient's tumor have done under half the dose?} The ground
truth re-simulates Equation~\ref{eq:tumor} for that patient with their
\emph{true} latent path, which is available only in simulation.

\paragraph{What the Figures Show.}
The left panel checks the matching: for four patients spanning the dose range it
plots the inferred rank trajectory (markers) against the rank trajectory of the
noise sample they were assigned (line). The two track each other, so each patient
is paired with a plausible sensitivity path whose ups and downs line up over
time. The center panel checks the fit at the population level, comparing observed
and generated trajectories as medians with $10$--$90\%$ bands; the
noise-conditioned operator reproduces both the typical trajectory and the spread
across patients. The right panel shows one counterfactual, for a high-dose
patient chosen for close agreement (the median counterfactual error across
patients is $0.57$ volume units): under half the dose the tumor regrows, and the
operator's prediction, computed from the \emph{inferred} noise, tracks the ground
truth computed from the \emph{true} noise.

\paragraph{Reading These Results.}
The counterfactuals here rest on strong assumptions, most importantly
monotonicity in the noise and a correctly specified mechanism. We present them as
a diagnostic for the model class rather than as claims about any real patient.

\end{document}